\documentclass[10pt,twocolumn,letterpaper]{article}

 \usepackage{iccv}              

\usepackage{bbding}
\usepackage{xcolor}
\usepackage{colortbl}
\usepackage{arydshln}

\definecolor{mygray}{gray}{.9}
\definecolor{hidden-red}{RGB}{205, 44, 36}
\definecolor{hidden-blue}{RGB}{194,232,247}
\definecolor{hidden-orange}{RGB}{243,202,120}
\definecolor{hidden-green}{RGB}{34,139,34}
\definecolor{hidden-pink}{RGB}{255,245,247}
\definecolor{hidden-black}{RGB}{20,68,106}
\definecolor{LightRed}{rgb}{1,0.92,0.92}
\definecolor{LightOrange}{rgb}{1,0.95,0.88}
\definecolor{LightYellow}{rgb}{1.0,1.0,0.84}
\definecolor{LightGreen}{rgb}{0.9,1.0,0.88}
\definecolor{LightCyan}{rgb}{0.9,1,1}
\definecolor{LightBlue}{rgb}{0.9,0.94,1}
\definecolor{LightIndigo}{rgb}{0.92,0.9,1}
\definecolor{LightMagenta}{rgb}{0.96,0.86,1}
\definecolor{DirtyWhite}{rgb}{0.96,0.96,0.96}
\definecolor{DRed}{RGB}{255,0,0}

\definecolor{iccvblue}{rgb}{0.21,0.49,0.74}
\usepackage[pagebackref,breaklinks,colorlinks,allcolors=iccvblue]{hyperref}

\def\paperID{5709} 
\def\confName{ICCV}
\def\confYear{2025}

\title{Interpretable Zero-Shot Learning with Locally-Aligned Vision-Language Model}

\author{Shiming Chen$^{1}$, Bowen Duan$^{2}$, Salman Khan$^{1,3}$, Fahad Shahbaz Khan$^{1,4}$\\
	$^{1}$Mohamed bin Zayed University of AI \quad
	$^{2}$Huazhong University of Science and Technology \\
	$^{3}$Australian National University \quad
	$^{4}$Linköping  University  \\
	\quad
	{\tt\small \{shimingchen, bwduan9910\}@gmail.com \quad \{salman.khan, fahad.khan\}@mbzuai.ac.ae}
}

\begin{document}
\maketitle

\begin{abstract}
	Large-scale vision-language models (VLMs), such as CLIP, have achieved remarkable success in zero-shot learning (ZSL) by leveraging large-scale visual-text pair datasets. However, these methods often lack interpretability, as they compute the similarity between an entire query image and the embedded category words, making it difficult to explain their predictions. One approach to address this issue is to develop interpretable models by integrating language, where classifiers are built using discrete attributes, similar to human perception. This introduces a new challenge: how to effectively align local visual features with corresponding attributes based on pre-trained VLMs. To tackle this, we propose LaZSL, a locally-aligned vision-language model for interpretable ZSL. LaZSL employs local visual-semantic alignment via optimal transport to perform interaction between visual regions and their associated attributes, facilitating effective alignment and providing interpretable similarity without the need for additional training. Extensive experiments demonstrate that our method offers several advantages, including enhanced interpretability, improved accuracy, and strong domain generalization. Codes available at: \url{https://github.com/shiming-chen/LaZSL}.
\end{abstract}

\section{Introduction}
\label{sec1}

Large-scale vision-language models (VLMs), such as CLIP \cite{RadfordKHRGASAM21},  have achieved significant zero-shot learning (ZSL) by training with large amount visual-text pairs, sparking a new wave of research in ZSL \cite{Palatucci2009ZeroshotLW,Xian2019ZeroShotLC}. Specifically, CLIP performs a zero-shot classification procedure by computing the similarity between the whole query image and the
embedded words for each category prompt (e.g.,  “\textit{a photo of a {class}}”), then choosing the highest similarity for classification. Since its encoder includes rich information in the world and does not require additional annotation knowledge, it is also used in segmentation \cite{Tang2024HuntingAC}, detection \cite{Kaul2023MultiModalCF}, and retrieval \cite{Du2024Image2SentenceBA} tasks.

Despite its achievements, CLIP’s performance exhibits notable sensitivity to the text prompts used during the inference stage and the domain-specific datasets. Accordingly, many works focus on i)  prompts learning \cite{ZhouYL022,ZhouYLL22, ShuNHYGAX22} and ii) adapter learning \cite{Gao2023CLIPAdapterBV, KhattakR0KK23,Murugesan2024RobustCO}. Prompt learning discovers the domain knowledge from the downstream data for improving the generalization of CLIP.  For example, Zhou introduced prompt learning using the downstream data for learning domain knowledge in the text prompt \cite{ZhouYL022, ZhouYLL22}. Shu \cite{ShuNHYGAX22} and Feng \cite{Feng0LKZ23} explored additional insights from the test
sample itself to enhance the domain knowledge of the prompt. Adapter learning follows fine-tuning thoughts to learn lightweight parameters via additional visual-semantic interactions \cite{KhattakR0KK23}. 

\begin{figure*}[t]
	\begin{center}
		\includegraphics[width=1\linewidth]{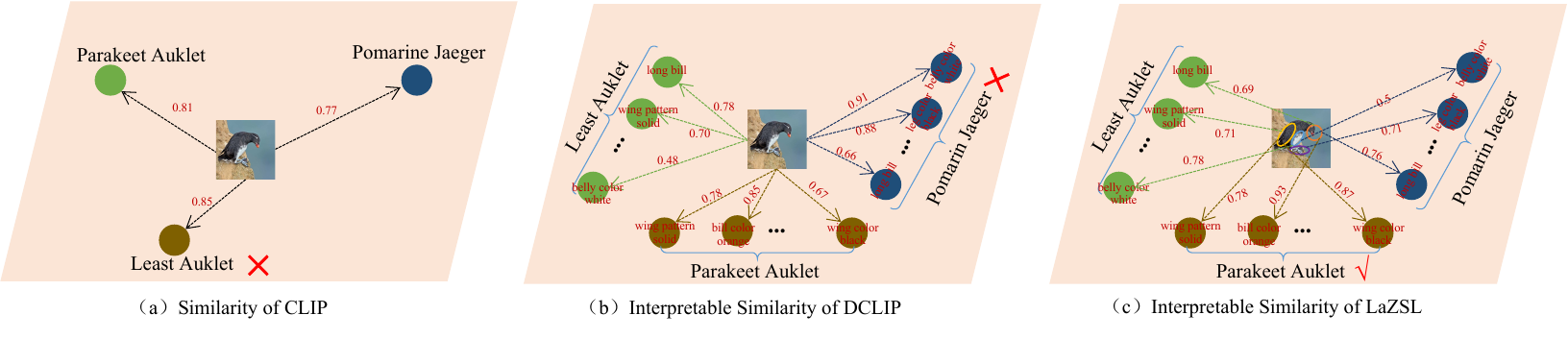}
		\\\vspace{-3mm}
		\caption{Comparison between the similarity of CLIP \cite{RadfordKHRGASAM21}, DCLIP \cite{Menon2022VisualCV},  and our LAZSL. (a) CLIP simply computes the similarity between the whole query image and the embedded words for each category,  failing to explain their predictions. (b) DCLIP  builds interpretable similarity based on alignments between whole query images and attributes, obtaining limited generalization.    (c) Our LaZSL represents the optimally interpretable similarity between local visual regions and attributes.} 
		\label{fig:paradigm}
	\end{center}
\end{figure*}

However, these methods compute the similarity between the whole query image and the embedded words for each category prompt following the standard CLIP, resulting in limited interpretability, as shown in Fig. \ref{fig:paradigm}(a). Because they cannot recognize classes based on the corresponding factors (e.g., attributes/semantics) of target classes. Meanwhile, they fail to capture fine-grained information of vision, their generalization is limited. Accordingly, a few works \cite{Menon2022VisualCV, Pratt2023WhatDA, Fan2023ImprovingCT,RothKKVSA23, Chiquier2024EvolvingIV,TianZYZ24} attempt to build interpretable models by integrating language, where classifiers are constructed with the discrete attributes. Specifically, they adopt
large language models (LLMs) to generate multiple finer text
descriptions with amounts of attribute for each category, and take these text as text prompts for computing similarity with image, as shown in Fig. \ref{fig:paradigm}(b).  They can classify classes according to the key attributes corresponding to the target classes, and thus their classifiers have interpretability analogously to human perception, where humans recognize an object via the obvious factors (e.g., attributes). Unfortunately, these interpretable similarities are between the whole image and attributes, which cannot directly capture the relationships between fine-grained visual information and their corresponding attributes, inevitably resulting in wrong visual-semantic alignments and limiting the generalization of VLMs.

Naturally, interpretable  ZSL raises a new challenge: how to conduct effective alignment between local vision and attributes based on the pre-trained VLMs. Different from classical ZSL methods \cite{0002HXYPWZY22,ChenHHXSZYYS23,Liu0ZWBZ23, NaeemXGT24} that can take attention mechanism in the network backbone to learn attribute localization for local alignment between visual regions and human annotated attributes, the network of VLMs are frozen and not easy to re-design for training as there are not sufficient data. Motivated by optimal transport (OT) theory \cite{0003LLX0Z23,LiWLZLCZ23,PeyreC19}, we can divide each image as a set of local patches over the visual space and view each attribute set of one class as
a discrete distribution over the semantic space. With such formulation, the classification task
becomes to measure the distance between the two distributions of visual and semantic spaces, as shown in Fig. \ref{fig:paradigm}(c). The plan of OT is then calculated between the local visual features and attribute features, enabling the local alignments. Accordingly,  it is possible to predict the label with the detailed attributes and patch features, resulting in more effective alignment and classification accuracy.

In this paper, we propose a locally-aligned vision-language model for interpretable ZSL, dubbed LaZSL. LaZSL first constructs the vision and semantic sets by randomly cropping the image and LLM, respectively.  Then, LaZSL adopts local visual-semantic alignment via OT to formulate the interaction between the constructed visual sets and semantic sets, enabling effective alignments by optimizing the OT plan. Notably, we also incorporate the global visual information into the hybrid cost matrix to avoid knowledge forgetting in the pre-trained VLMs. Finally, we can predict their classes of input images by aligning hybrid similarity matrix with OT plan. The extensive experiments on nine widely-used data demonstrate our method offers several advantages, i.e., interpretability, improvements in accuracy, and good domain generalization.

Our main contributions are summarized in the following:
\begin{itemize}
	\item We propose LaZSL to conduct effective alignment between local vision and attributes based on the pre-trained VLMs (e.g., CLIP) for interpretable ZSL. Different from most variants of CLIP that require additional model training, our LaZSL is training-free.
	
	\item We introduce local visual-semantic alignment using optimal transport to formulate interaction between the constructed visual regions and attributes, enabling effective alignment and obtaining interpretable similarity for ZSL prediction.
	
	\item We conduct extensive experiments on nine widely-used datasets to evaluate our methods, and results demonstrate that our LaZSL achieves competitive performances over baselines.
\end{itemize}

\section{Related Works}
\label{sec2}

\noindent \textbf{Classical Zero-Shot Learning:} Early ZSL methods utilize the human annotated attributes as side-information for knowledge transfer from seen classes to unseen ones \cite{Palatucci2009ZeroshotLW,Xian2019ZeroShotLC}.  They target how to conduct effective visual-semantic interactions for knowledge transfer. Typically, there are three types of interactions: embedding-based methods, generative methods, and common space learning methods. Embedding-based methods map visual features into semantic space and search nearest-neighbor semantic prototypes for classification \cite{AkataPHS16,HuynhE20,ChenHHXSZYYS23,ChenH0K24}. Generative methods learn a semantic-conditioned generator to synthesize image/feature samples for unseen classes and transform the ZSL task as supervised classification \cite{Xian2018FeatureGN,ChenS25,Chen2023EvolvingSP, ShimingGenZSL25}.  Common-space learning methods map the visual and semantic features into a common space and perform classification by nearest-neighbor search \cite{WangC17,SchonfeldESDA19,Chen2021HSVA}. Although these methods have interpretability with the human-annotated attributes, they are time-consuming and labor-intensive to collect for various scene generations. Thus, they cannot work well on the large-scale dataset, e.g., ImageNet.
\begin{figure*}[ht]
	\begin{center}
		\includegraphics[width=1\linewidth]{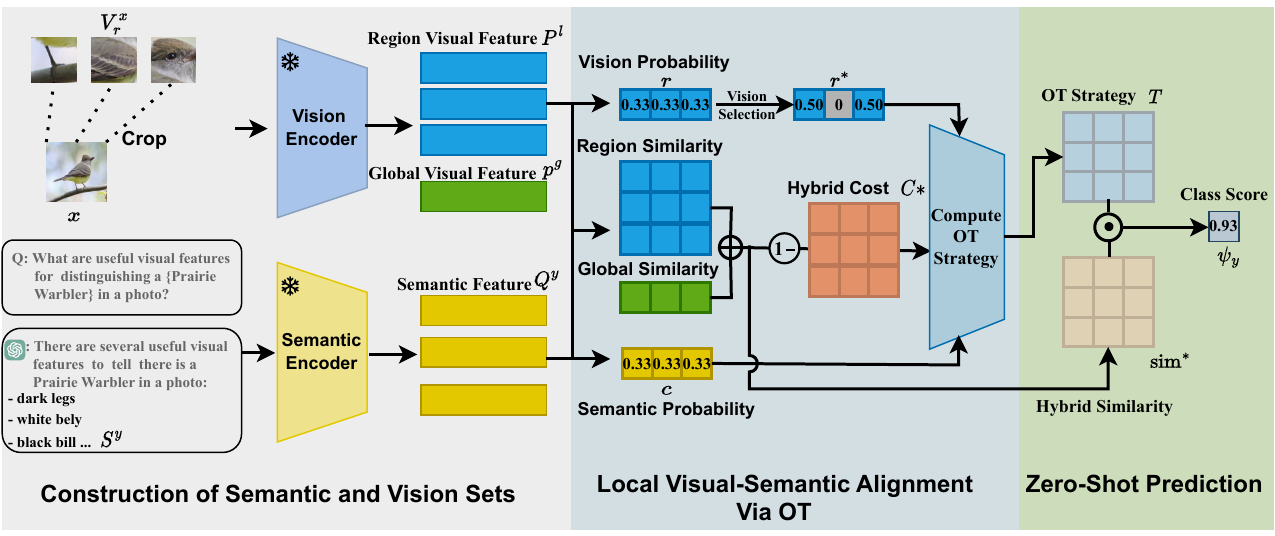}
		\\
		\caption{The pipeline of our LaZSL. LaZSL consists of three main module components, including the construction of semantic and vision sets, local visual-semantic alignment via OT, and zero-shot prediction.} 
		\label{fig:pipeline}
	\end{center}
\end{figure*}

\noindent \textbf{VLM-Based Zero-Shot Learning:} VLMs take large-scale visual-text pairs for model training and have significant knowledge transfer capacity \cite{RadfordKHRGASAM21, JiaYXCPPLSLD21}. For example, CLIP leads new trends in ZSL tasks \cite{RadfordKHRGASAM21}. Despite its advances, CLIP heavily relies on prompt design and obtains limited performances in the domain-specific datasets \cite{ZhouYL022, 0004LEF0L24}. Many works attempt to improve CLIP by prompt learning \cite{ZhouYL022, KhattakR0KK23, ZhangKSNCK23}, adapter learning \cite{KhattakR0KK23,Murugesan2024RobustCO}. However, these methods compute the similarity between the whole query images and the word embedding of class names, which lacks interpretability. Recently, a few works \cite{Menon2022VisualCV, Pratt2023WhatDA, Fan2023ImprovingCT, RothKKVSA23, Chiquier2024EvolvingIV,TianZYZ24} build interpretable models by integrating language generated by LLMs, where classifiers can recognize target classes according to whether the object has corresponding key attributes analogously to human conception. However, they cannot well capture the relationships of local visual information with the corresponding attributes, resulting in wrong alignments between visual and semantic space and hampers the generalization capacity of VLMs. To tackle this challenge, we design a locally-aligned vision-language model to conduct effective alignment between local vision and attributes based on the pre-trained VLMs.

\noindent \textbf{Optimal Transport in Vision:}
The Optimal Transport (OT) theory is initially introduced to solve the problem of how to reduce the cost when moving several items simultaneously \cite{PeyreC19}. Thanks to the property of distribution matching, OT has been widely applied in many computer vision tasks \cite{ArjovskyCB17,LaclauRMBB17, KolkinSS19}. More related to our works,  some works \cite{0003LLX0Z23,LiWLZLCZ23, 0002YSLR023} take OT  for prompt tuning. For example, Chen \textit{et al.}\cite{0002YSLR023} and Li \textit{et al.}\cite{LiWLZLCZ23} utilize OT to learn single- or multi-mode prompts at the same time to optimize text prompts. Differently, we take OT to find better alignments between the local visual sets and attribute sets by local alignment and thus obtain accurate similarity for classification.

\section{Locally-Aligned Vision-Language Model}
\label{sec3}

In this section, we introduce our locally-aligned vision-language model for interpretable zero-shot learning (dubbed LaZSL), which performs effective alignment between local vision and attributes by optimizing the interactions of semantic and visual features using OT. LaZSL can accurately capture the local fine-grained visual information according to its corresponding attributes for class prediction. Thus, it achieves interpretability, accuracy improvements, and good domain generalization. As shown in Fig. \ref{fig:pipeline}, LaZSL consists of three main components: the construction of semantic and visual sets, the visual-semantic interaction at the attribute level based on OT, and the zero-shot prediction. We will introduce their details in the following subsections.

\subsection{Construction of Semantic and Vision Sets}
Although CLIP variants based on global features can handle ZSL tasks effectively, they often struggle with fine-grained classification or categories with semantically sparse names. Accordingly, enriching CLIP models with local attributes to capture fine-grained visual information is necessary. Furthermore, formulating an attribute classifier has the property of interpretability \cite{Menon2022VisualCV}. Targeting this goal, we first construct the semantic and vision sets.

\noindent\textbf{Construction of Semantic Sets.} To introduce semantic information at the attribute level, we follow \cite{Menon2022VisualCV} to construct a semantic set for each category with the help of LLMs. Briefly, given a class label $y$, the semantic sets $S^y$ can be generated:
\begin{equation}
	\label{eq:1}
	S^y=h(prompt(y))=\{s^y_{i}|i=1,\dots, M\},
\end{equation}
where $h(\cdot)$ is an LLM (e.g., GPT-3) and $prompt(\cdot)$ is a template function that generates queries for LLMs by utilizing fixed prompts.

In fact, after obtaining the semantic set for each category, most interpretable ZSL methods simply compute the category score by averaging the similarity between global visual features and the attributes in the semantic set \cite{Menon2022VisualCV,Pratt2023WhatDA,RothKKVSA23}. While this approach can partially alleviate the deficiencies caused by missing attribute semantics, it essentially performs global alignment with attribute information rather than achieving true attribute-level visual-semantic interaction. To address this, we further construct a visual set to capture local visual information, which is then used for local visual-semantic interaction.

\noindent\textbf{Construction of Vision Sets.} Specifically, we attempt to capture visual regions corresponding to the semantic set by performing random multi-scale cropping on the original images.
Given an image $x \in \mathbb{R}^{H \times W \times 3}$, where H and W are its height and width respectively, we propose $P_r(\cdot,\cdot)$ to construct region vision sets:
\begin{equation}
	\label{eq:2}
	V_r^x=\{v_i^x=P_r(x,\gamma_i min(W,H))|i=1,\dots,N\},
\end{equation}
where \( \gamma_i \) is a random number drawn from the uniform distribution \( U(\alpha, \beta) \), where \( \alpha \) and \( \beta \) are predefined hyperparameters that constrain the lower and upper bounds of the sampling range respectively. $N$ is the number of cropped regions and typically set to $[60-90]$. The function \( P_r(\cdot, \cdot) \) performs a random region cropping on an image, with \( \gamma_i \)  specifying the scale of the cropped regions. It can be seen that \( P_r(\cdot, \cdot) \) and \( \gamma_i \) ensure that the constructed region vision set contains visual features of various scales. This enables our visual set to better match the previously constructed attribute set.

\subsection{Local Visual-Semantic Alignment Via OT}\label{3.2}
After constructing the visual and semantic sets, we use CLIP encoders to obtain their latent space representations $P_x=\{p_i\}_{i=1}^N$ and $Q_y=\{q_j\}_{j=1}^M$,  respectively:
\begin{align}
	\label{eq:4}& P^x=P^l \cup \{p^g\} =\sigma_v(V_r^x) \cup \{\sigma_v(x)\},\\
	\label{eq:5}& Q^y=\sigma_s(S^y),  
\end{align}
where, $\sigma_v$ and $\sigma_s$ are the vision encoder and semantic encoder from CLIP,respectively.

Unlike existing interpretable ZSL methods \cite{Menon2022VisualCV,Pratt2023WhatDA,RothKKVSA23} that perform visual-semantic alignment using global features (1 to 1), our approach constructs local visual and semantic sets (N to M). Simply averaging multiple similarities is insufficient to fully estimate the similarity between the two sets. Therefore, we propose an OT matching method. 

The optimal transport problem can be viewed as finding the minimum cost to transform one probability distribution into another. Since this cost can serve as a measure of the distance between two probability distributions, we employ OT theory to match the vision set and the semantic set. The process to compute OT is formulated as:
\begin{align}
	\label{eq:6-1}& T=\mathrm{OT}(C,r,c),
\end{align}
where T is the OT plan, $r$ and $c$ are the discrete probability vectors belonging to the vision set and semantic set respectively, and can be initialized as uniform distributions. $C$  is the cost matrix between the region vision set and the semantics set, and $C_{i,j}=1-\mathrm{sim}_{i,j}=1-\mathrm{cosine}(p^l_i,q^y_j)$

It can be seen that, by using the OT, we have found an effective way to evaluate the similarity between the vision set and the semantic set. However, this algorithm still has two limitations: i) the randomly obtained region visual features contain some noises, and ii) the region visual features may cause the knowledge forgetting of CLIP visual encoder and cannot well capture their corresponding categories. To address these issues, we propose a vision selection mechanism and a region-global hybrid cost approach.

\noindent\textbf{Vision Selection.} Specifically, the original region visual features are divided into a relevant region set and an irrelevant region set based on a threshold $\delta$, which is obtained according to the average similarity of all region visions and the global image. Then, we remove irrelevant noises by modifying the initialization of \( r \) that is relevant to $\delta$. This process can be expressed as:
\begin{align}
	\label{eq:10}& \delta=\mathrm{AVG}(\sum_{i=1}^N{\mathrm{cosine}(p^l_i,p^g)}), \\
	\label{eq:11}& P^l_{pos}=\{p^l_i|\mathrm{cosine}(p^l_i,p^g) \geq \delta \}, \\
	\label{eq:12}& P^l_{neg}=\{p^l_i|\mathrm{cosine}(p^l_i,p^g) < \delta \}, \\
	\label{eq:13}& r_i^* = 
	\begin{cases}
		\frac{1}{|P^l_{pos}|} & \mathrm{if}\ p^l_i \in P^l_{pos},\\ 
		0 & \mathrm{otherwise}.
	\end{cases}
\end{align}
Thus, the visual probability $r$ is updated as $r^*$ according to Eq. \ref{eq:13}. Then, the positive region vision set $p_i^l\in P_{pos}^l$ is used for visual-semantic interaction via OT.

\noindent\textbf{Region-Global Hybrid Cost.} 
In OT, the cost matrix is a crucial source of prior knowledge. In the visual-semantic interaction based on OT, we observe that our cost is solely composed of the similarity between the randomly cropped region vision set and the semantic set. This inevitably makes the OT strategy $T$, computed from the cost, overly sensitive to the noise introduced by the cropping process, resulting in knowledge forgetting of the vision encoder in CLIP. Therefore, we choose to incorporate additional global prior information into the cost matrix to address this issue:
\begin{equation}
	\label{eq:14}
	C^*_{i}=1-(\theta \mathrm{sim}_i+(1-\theta)p^{g\top}Q^y)
\end{equation}
where $C^*_{i}$ is the $i$-th row of the hybrid cost matrix $C^*$, $\theta \in (0,1) $ is a hyper-parameter, which can be seen as the hybrid confidence between region feature and global feature.

\noindent\textbf{Compute OT Strategy.} 
Once we have obtained $C^*$, we can compute the optimal transport plan $ T $. Specifically, we choose the Sinkhorn algorithm \cite{Cuturi13} to solve the optimal transport distance, which uses an entropic constraint for fast optimization. By iteratively updating the strategy matrix between the region vision set and semantic set,  Eq. \ref{eq:6-1} is specifically formulated as:
\begin{align}
	\label{eq:6}& T=\mathrm{diag}(\mathcal{U}) \mathcal{M} \mathrm{diag}(\mathcal{V}), \\
	\label{eq:7}& u_i^{k+1}=\frac{r^*_i}{\sum_j{\mathcal{M}_{i,j}v_j^k}}, \\
	\label{eq:8}& v_j^{k+1}=\frac{c_j}{\sum_i{\mathcal{M}_{i,j}u_i^k}},\\
	\label{eq:9}& \mathcal{M}=\mathrm{exp}(\frac{-C^*}{\lambda}),  
\end{align}
where T is the OT plan, $r^*$ and $c$ are the discrete probability vectors belonging to the vision set and semantic set. $C^*$  is the hybrid cost matrix between the region vision set and the semantics set, $k$ is the number of iterations for $T$, $\mathrm{diag}(\cdot)$ is a function used to construct a diagonal matrix, $\mathcal{U}=\{u_i\}_{i=1}^N$ and $\mathcal{V}=\{v_j\}_{j=1}^M$.

\subsection{Zero-Shot Prediction}
\label{3.3}
After the visual-semantic interaction, we get the OT plan $T$. Similarly, to align with the cost matrix, we calculate the category scores using a hybrid similarity approach:

\begin{align}
	\label{eq:15}&\psi_y=<T,\mathrm{sim}^*>_F,\\
	\label{eq:16}&\mathrm{sim}^*_i=\theta \mathrm{sim}_i+(1-\theta)p^{g\top}Q^y,
\end{align}
where $\mathrm{sim}^*_{i}$ is the $i$-th row of the hybrid similarity matrix $\mathrm{sim}^*$, $<\cdot,\cdot>_F$ is the Frobenius inner product between two matrix. It can be observed that for each category \( y \), we can obtain its corresponding category score. Next, we use this score to perform zero-shot predictions:
\begin{equation}
	y^*=\mathop{\arg\max}\limits_{y}\psi_y,
\end{equation}
where $y^*$ represents the predicted category label.

\section{Experiments}\label{sec4}
\subsection{Experiment Setup}
\noindent\textbf{Datasets.} To make a comprehensive evaluation of our work, we conduct extensive ZSL experiments on cross-dataset transfer
learning and domain generalization. These experiments are evaluated on nine widely used image
datasets, varying in scale and domains. For example, ImageNet \cite{DengDSLL009} is used for recognizing daily objects; CUB  \cite{Welinder2010CaltechUCSDB2} is used for fine-grained classification of birds, Oxford Pets \cite{ParkhiVZJ12} is used for recognizing common animals, Food101 \cite{BossardGG14} is specifically designed for food classification,  Place365 \cite{ZhouLKO018} is applied to scene recognition. We also take the variant datasets of ImageNet with natural domain shifts to validate the domain generalization capacity of LaZSL, including,
ImageNet-V2 \cite{RechtRSS19}, ImageNet-Sketch \cite{WangGLX19}, ImageNet-A \cite{HendrycksZBSS21} and
ImageNet-R \cite{HendrycksBMKWDD21}, each dataset represents a unique
distribution shift from ImageNet.

\noindent\textbf{Compared Baselines.} Since our method is VLM-based ZSL, we primarily take VLM-based ZSL methods for fair comparison on various network backbones, i.e., CLIP models with ViT-B/32, ViT-B/16, ViT-L/14. Compared baselines includes the standard VLM-based ZSL methods (i.e., CLIP~\cite{RadfordKHRGASAM21}, CoOp \cite{ZhouYLL22}, CoCoOp \cite{ZhouYL022}, TPT~\cite{ShuNHYGAX22},  MaPLe~\cite{KhattakR0KK23}, ProGrad~\cite{ZhuNHWZ23}) and interpretable VLM-based ZSL methods (i.e., DCLIP~\cite{Menon2022VisualCV}, WaffleCLIP~\cite{RothKKVSA23}, CuPL~\cite{Pratt2023WhatDA}, ArGue \cite{TianZYZ24}).

\noindent\textbf{Implementation Details.}
In this work, we mainly follow \cite{Menon2022VisualCV} to take GPT-3 to generate the attribute sets for each class. Specifically, we take the following prompts to generate attribute descriptions for each class:\\
\textit{\color{gray}Q: What are useful features for distinguishing a \{class name\} in a photo?}\\
\textit{\color{gray}A: There are several useful visual features to tell there is a \{class name\}  in a photo:}\\
{\color{gray}--}\\
To ensure more consistent output from LLM, we additionally included two specific examples in our prompt. The text prompt template of text encoder for classification is “\textit{\color{gray}A photo of a \{class name\}, which (is/has/etc)
	\{attribute\}}”. During visual sets construction, we set the cropping scale to 0.6 and each image is typically cropped into $[60, 90]$ region patches. The hybrid coefficient $\theta$ is set to 0.8 for all datasets. 
All experiments are performed on a single NVIDIA H100 graphic card with 80GB memory.

\begin{table*}[ht]
	\small
	\centering  
	\caption{Comparison of ZSL performance (accuracy in \%) across five image classification benchmarks (i.e., ImageNet, CUB,  Oxford Pets,  Food101, and  Place365) using three different CLIP models (ViT-B/32, ViT-B/16, ViT-L/14).  {\color{red}$\bigtriangleup$ DCLIP} denotes the improvements of our LaZSL over DCLIP. For each group, the best results are marked with \textbf{Bold}.}\label{table:sota}
	\resizebox{1.0\linewidth}{!}{\small
		\begin{tabular}{l|c|c|c|c|c|c}
			\hline
			Method &ImageNet& CUB&  Oxford Pets &  Food101 &  Place365 &Average\\
			\hline
			&\multicolumn{6}{c}{\textit{ViT-B / 32 with pre-trained weights from CLIP}}\\
			\cline{2-7}
			CLIP~\cite{RadfordKHRGASAM21}&62.1&51.2& 85.0& 82.6& 38.5&63.3\\
			DCLIP~\cite{Menon2022VisualCV}&63.0&52.7& 84.5& 84.1& 39.9&64.8\\
			WaffleCLIP~\cite{RothKKVSA23}& 63.3& 52.0& 85.5& 84.0&39.5&64.9\\
			CuPL~\cite{Pratt2023WhatDA}& 64.4& 49.8&87.0& 84.2&39.1&64.9\\
			ProAPO \cite{QuProAPO25} &64.1&53.6&\textbf{88.7}&84.2&\textbf{42.7}&66.7\\
			\rowcolor{LightBlue}\textbf{LaZSL (Ours)}&\textbf{65.3}&	\textbf{56.5}& 84.7 &\textbf{85.9}&41.5&\textbf{66.8}\\
			\hdashline
			\rowcolor{LightBlue}{\color{red}$\bigtriangleup$ DCLIP}&{\color{red} +2.2}&{\color{red} +3.8}&{\color{red} +0.2}&{\color{red} +1.8}&{\color{red} +1.6}&{\color{red} +2.0}\\
			\hline
			&\multicolumn{6}{c}{\textit{ViT-B / 16 with pre-trained weights from CLIP}}\\
			\cline{2-7}
			CLIP~\cite{RadfordKHRGASAM21}&66.7&56.0& 88.1& 88.4& 39.3&67.7\\
			DCLIP~\cite{Menon2022VisualCV}&67.9&57.1& 86.9& 88.5& 40.3&68.1\\
			WaffleCLIP~\cite{RothKKVSA23}& 68.1& 56.9& 86.5& 89.1&40.8&68.3\\
			CuPL~\cite{Pratt2023WhatDA}& \textbf{69.6}& 56.4& \textbf{91.1}& 89.0&39.8&69.2\\
			\rowcolor{LightBlue}\textbf{LaZSL (Ours)}&69.2&	\textbf{60.3}& 87.4 &\textbf{89.7}&\textbf{42.0}&\textbf{69.7}\\
			\hdashline
			\rowcolor{LightBlue}{\color{red}$\bigtriangleup$ DCLIP}&{\color{red} +1.3}&{\color{red} +3.1}&{\color{red} +0.52}&{\color{red} +1.2}&{\color{red} +1.7}&{\color{red} +1.6}\\
			\hline	
			&\multicolumn{6}{c}{\textit{ViT-L / 14 with pre-trained weights from CLIP}}\\
			\cline{2-7}
			CLIP~\cite{RadfordKHRGASAM21}&73.5&62.1& 93.2&  92.6& 39.6&72.2\\
			DCLIP~\cite{Menon2022VisualCV}&74.9&63.5& 92.4&  93.0& 40.3&72.8\\
			WaffleCLIP~\cite{RothKKVSA23}& 75.3& 62.3& 91.6& 93.3&40.9&72.7\\
			CuPL~\cite{Pratt2023WhatDA}& \textbf{76.6}& 62.2& \textbf{94.3}& 93.4&40.8&73.5\\
			\rowcolor{LightBlue}\textbf{LaZSL (Ours)}&75.7&	\textbf{66.1}& 92.7 &\textbf{93.5}&\textbf{41.8}&\textbf{74.0}\\
			\hdashline
			\rowcolor{LightBlue}{\color{red}$\bigtriangleup$ DCLIP}&{\color{red} +0.8}&{\color{red} +2.6}&{\color{red} +0.3}&{\color{red} +0.5}&{\color{red} +1.5}&{\color{red} +1.2}\\
			\hline	
	\end{tabular} }
\end{table*}

\subsection{Comparison with State of the Arts }
\noindent \textbf{Results on Cross Datasets. } We first compared our LaZSL with the interpretable ZSL methods (i.e., DCLIP~\cite{Menon2022VisualCV}, WaffleCLIP~\cite{RothKKVSA23}, CuPL~\cite{Pratt2023WhatDA}) on five image classification datasets. Results are shown in Table \ref{table:sota}. We find that LaZSL achieves the best average performances across five datasets when using various network backbones, i.e., 66.8\%, 69.7\%, and 74.0\% average accuracy on ViT-B/32, ViT-B/16, and  ViT-L/14, respectively. Compared to DCLIP \cite{Menon2022VisualCV} that is the first interpretable ZSL method based on VLM, LaZSL consistently obtains performance gains on all datasets with various network backbones. Especially, the performance gains are larger on the challenging fine-grained datasets (e.g., CUB, Place365). Because aligning key fine-grained visual details is important for the alignment of vision and attributes. These results reveal that LaZSL is effective in conducting alignment between local vision and attributes based on the pre-trained VLM for interpretable ZSL.

\noindent \textbf{Results on Cross Domains. } We also evaluate the domain generalization capacity of LaZSL on the variants of ImageNet with natural domain shifts, as shown in Table \ref{table:generalization}.  Compared to the standard VLM-based methods (e.g.,  CoOp \cite{ZhouYLL22}, CoCoOp \cite{ZhouYL022}, TPT~\cite{ShuNHYGAX22},  MaPLe~\cite{KhattakR0KK23}, ProGrad~\cite{ZhuNHWZ23}) that require additional training for prompt or models, LaZSL advances the interpretable ZSL methods to obtain state-of-the-art average performance (i.e., 60.9\%) over all datasets. This reveals that the interpretable classifier with attributes has great potential for ZSL by effective local alignment without additional training. When comparing with other interpretable  ZSL methods, LaZSL achieves performance gains of 1.7\% and 0.9\% over DCLIP~\cite{Menon2022VisualCV} and CuPL~\cite{Pratt2023WhatDA}, respectively.  Notably, LaZSL even outperforms ArGue \cite{TianZYZ24} which is interpretable ZSL method tuning prompts with additional training. These should be thanks to the local visual-semantic alignment via OT, enabling to learn interpretable classifiers that can accurately capture the local visual information corresponding to their attributes.

\begin{table*}[ht]
	\small
	\centering  
	\caption{ ZSL performances (accuracy in \%) on ImageNet  with natural distribution shifts of the state-of-the-art methods. ViT-B/16 is used as a network backbone for fair comparison. Methods take the source data for training marked with {\color{red}\CheckmarkBold}, otherwise, marked with {\color{red} \XSolidBrush}. $\dagger$ denotes interpretable ZSL with attributes generated by LLMs. For each group, the best results are marked with \textbf{Bold}. } \label{table:generalization}
	\resizebox{1\linewidth}{!}{\small
		\begin{tabular}{l|c|c|c|c|c|c}
			\hline
			Method &Training&ImageNet-V2& ImageNet-R& ImageNet-S& ImageNet-A& Average\\
			\hline
			CLIP \cite{RadfordKHRGASAM21} &{\color{red}\XSolidBrush}&60.8& 74.0& 47.8 &46.1& 57.2\\
			CoOp \cite{ZhouYLL22}&{\color{red}\CheckmarkBold}& 64.2& 75.2& 47.9& 49.7& 59.3\\
			CoCoOp \cite{ZhouYL022}&{\color{red}\CheckmarkBold}&64.1& 76.2& 48.8& 50.6& 59.9\\
			TPT~\cite{ShuNHYGAX22}& {\color{red}\CheckmarkBold}&64.3& 73.9& 46.4& 53.6& 59.5\\
			MaPLe~\cite{KhattakR0KK23}& {\color{red}\CheckmarkBold}&64.1& 77.0& \textbf{49.2}& 50.9&60.3\\ 
			ProGrad~\cite{ZhuNHWZ23} &{\color{red}\CheckmarkBold}&\textbf{64.7}& 74.5& 47.9& 49.3& 59.1\\
			DCLIP$^\dagger$~\cite{Menon2022VisualCV}& {\color{red}\XSolidBrush}& 61.6&75.0&47.1&49.2&58.2\\
			CuPL$^\dagger$~\cite{PrattCLF23}& {\color{red}\XSolidBrush}& 63.3& \textbf{77.1}& 48.8& 50.8&60.0\\
			SaLS \cite{MurugesanSAD24} &{\color{red}\CheckmarkBold}&64.2&74.4&46.3& 48.5&58.4\\
			ArGue$^\dagger$ \cite{TianZYZ24}&{\color{red}\CheckmarkBold}&64.6&76.6&48.9&50.9&60.3\\
			\rowcolor{LightBlue}\textbf{LaZSL$^\dagger$ (Ours)}&{\color{red}\XSolidBrush}&63.3&	75.6& 48.2 &\textbf{56.2}&\textbf{60.9}\\
			\hline
	\end{tabular} }
\end{table*}

\begin{table}[t]
	\centering
	\small
	\caption{ Ablation studies for different components of our LaZSL on three datasets.  “OT” denotes local visual-semantic alignment via OT,  “VS” denotes vision selection,  “Hybrid” denotes hybrid the local and global visual information. We conduct these experiments using the CLIP model of ViT-B/16.  } \label{Table:ablation-components}
	\resizebox{1\linewidth}{!}
	{\small
		\begin{tabular}{l|ccc}
			\hline
			Method & ImageNet& CUB &Place365\\
			\hline  
			Baseline   & 67.9&57.8&40.3\\                       
			Baseline+OT &68.5& 59.0&	41.6\\   
			Baseline+OT+VS &69.0 &60.0& 41.8\\     
			Baseline+OT+Hybrid&69.0&	59.3 &41.9\\ 
			LaZSL (full)&\textbf{69.2 }&\textbf{60.3}& \textbf{42.0}\\   
			\hline 
		\end{tabular}
	}
\end{table}
\subsection{Ablation Study}
To provide further insight into LaZSL, we conduct ablation studies to evaluate the effects of various model components, including locally visual-semantic alignment via OT (denoted as “OT”), vision selection (denoted as “VS”), and hybrid the local and global visual information (denoted as “Hybrid”). Results on three datasets are presented in  Table \ref{Table:ablation-components}. The baseline is the typical interpretable ZSL method DCLIP \cite{Menon2022VisualCV}. When baseline using OT to conduct local visual-semantic alignment, its performances consistently improved by 0.6\%, 1.2\%, and 1.3\% on ImageNet, CUB, and Place365, respectively. This demonstrates it's necessary to perform local alignment in interpretable ZSL and our method can effectively advance it.  Furthermore, the visual selection and hybrid local and global visual information can further enable LaZSL to conduct accurate local visual regions with their corresponding attributes. Thus,  LaZSL can learn an accurate and interpretable classifier for ZSL based on the pre-trained VLM.

\begin{figure}
	\includegraphics[width=1\linewidth]{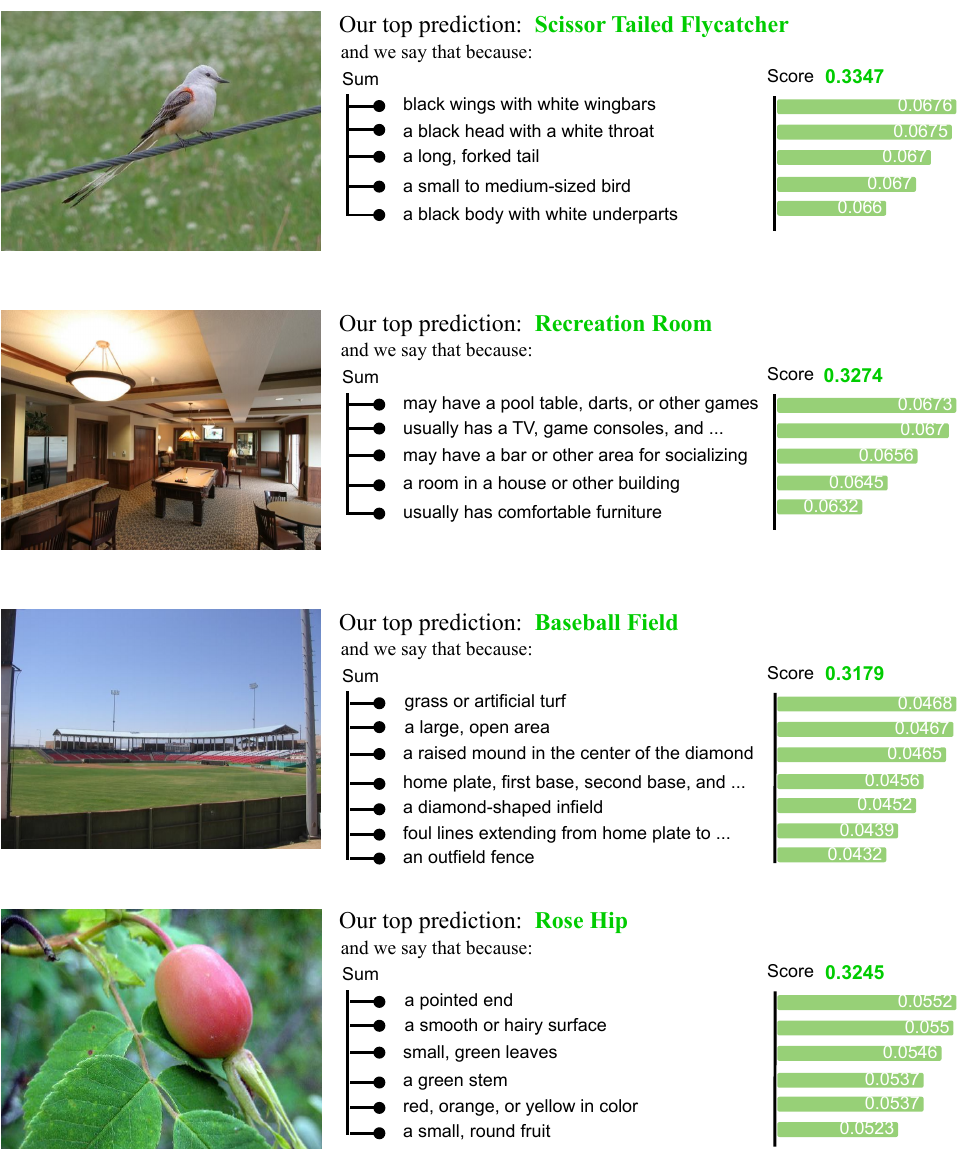}\\\vspace{-3mm}
	\caption{Visualization of interpretable classification of LaZSL.}
	\label{fig:visualization1}
\end{figure}

\begin{figure*}[htbp]
	\centering
	\includegraphics[width=1\linewidth]{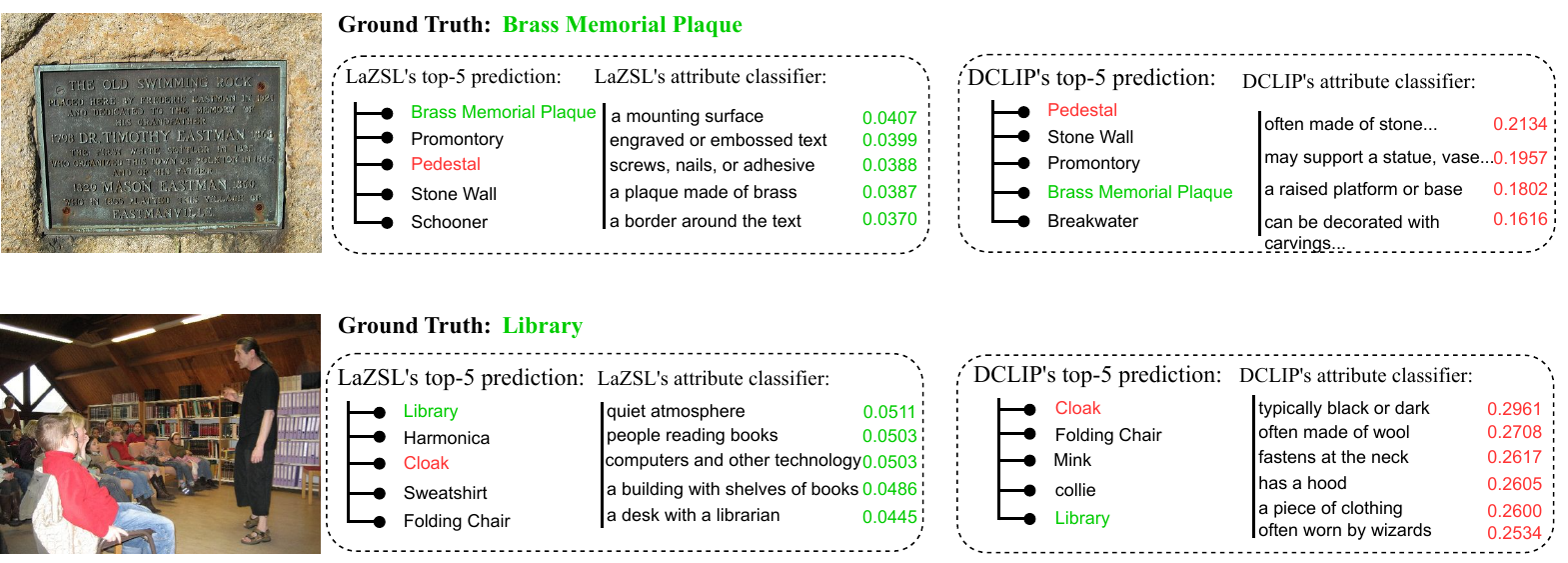}\\ \vspace{-3mm}
	\caption{Classification comparison between LaZSL and DCLIP \cite{Menon2022VisualCV}.}
	\label{fig:visualization2}
\end{figure*}

\renewcommand{\tabcolsep}{1pt}
\begin{figure*}[htbp]
	\begin{center}
		\vspace{-4mm}
		\begin{tabular}{cccc}
			\includegraphics[width=0.25\linewidth]{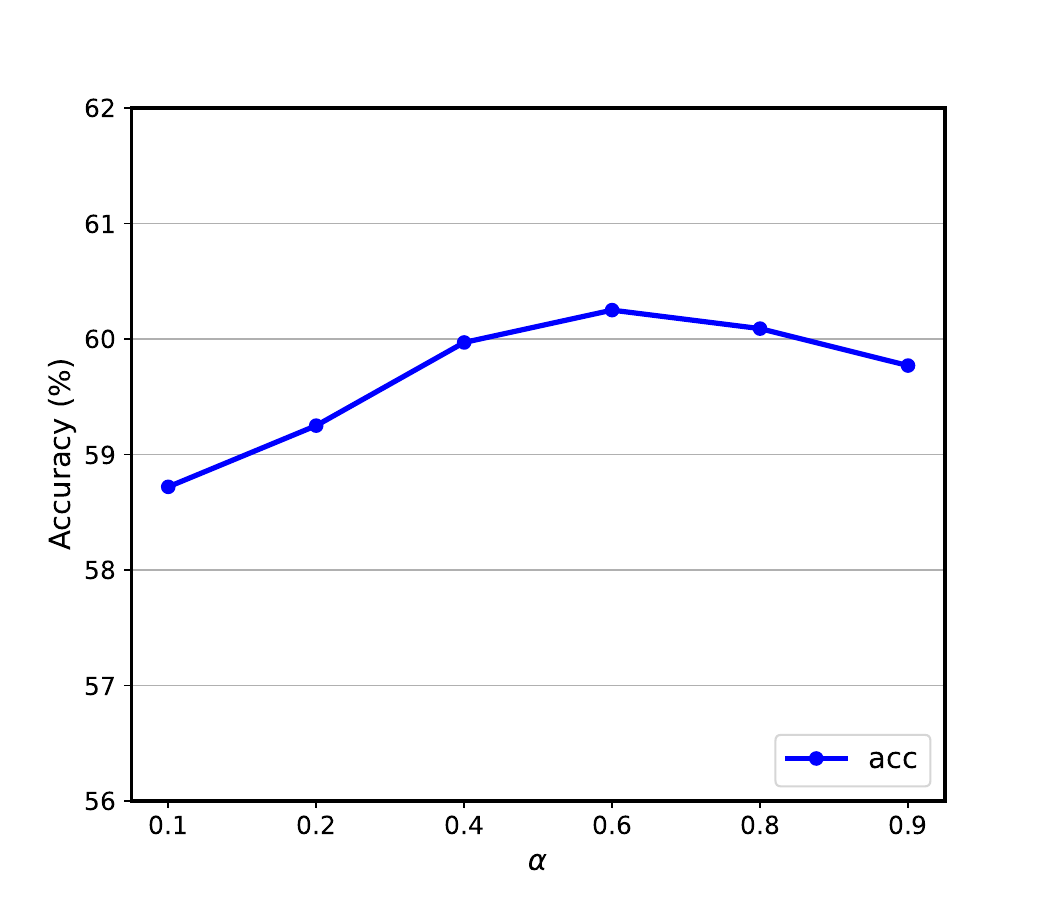}&
			\includegraphics[width=0.25\linewidth]{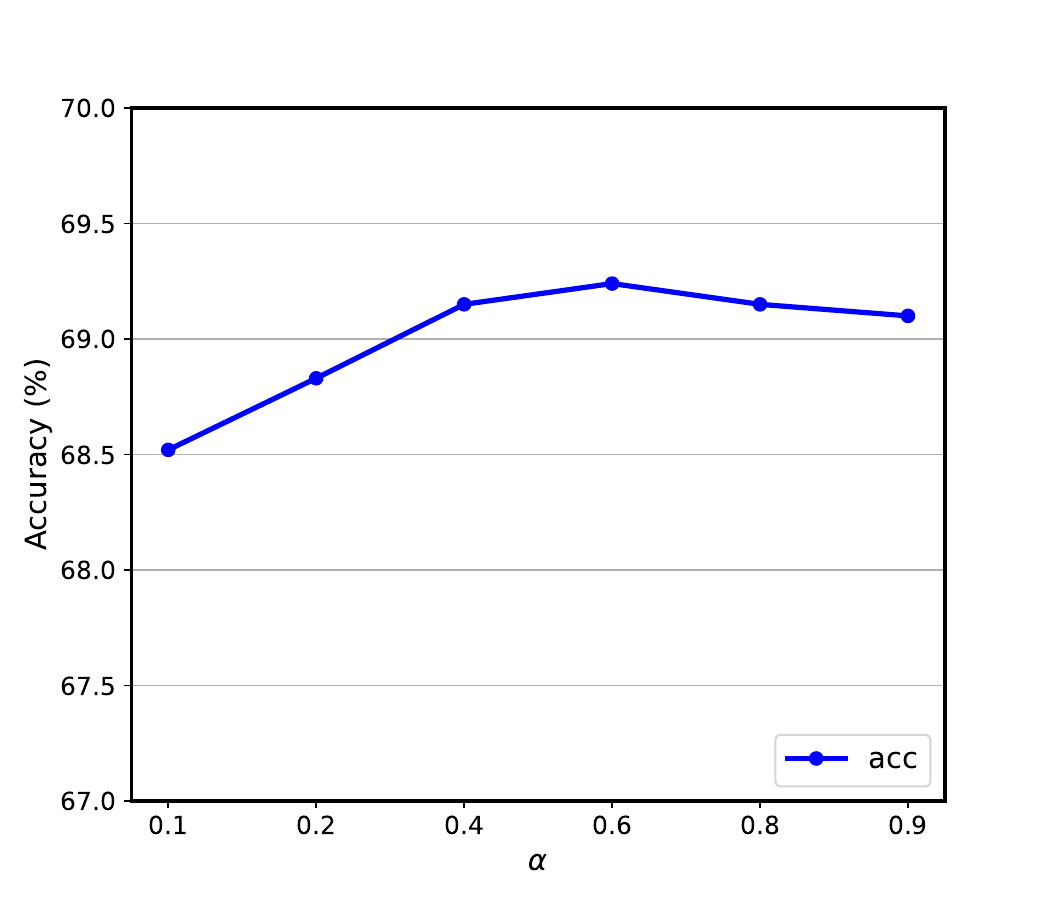}&
			\includegraphics[width=0.25\linewidth]{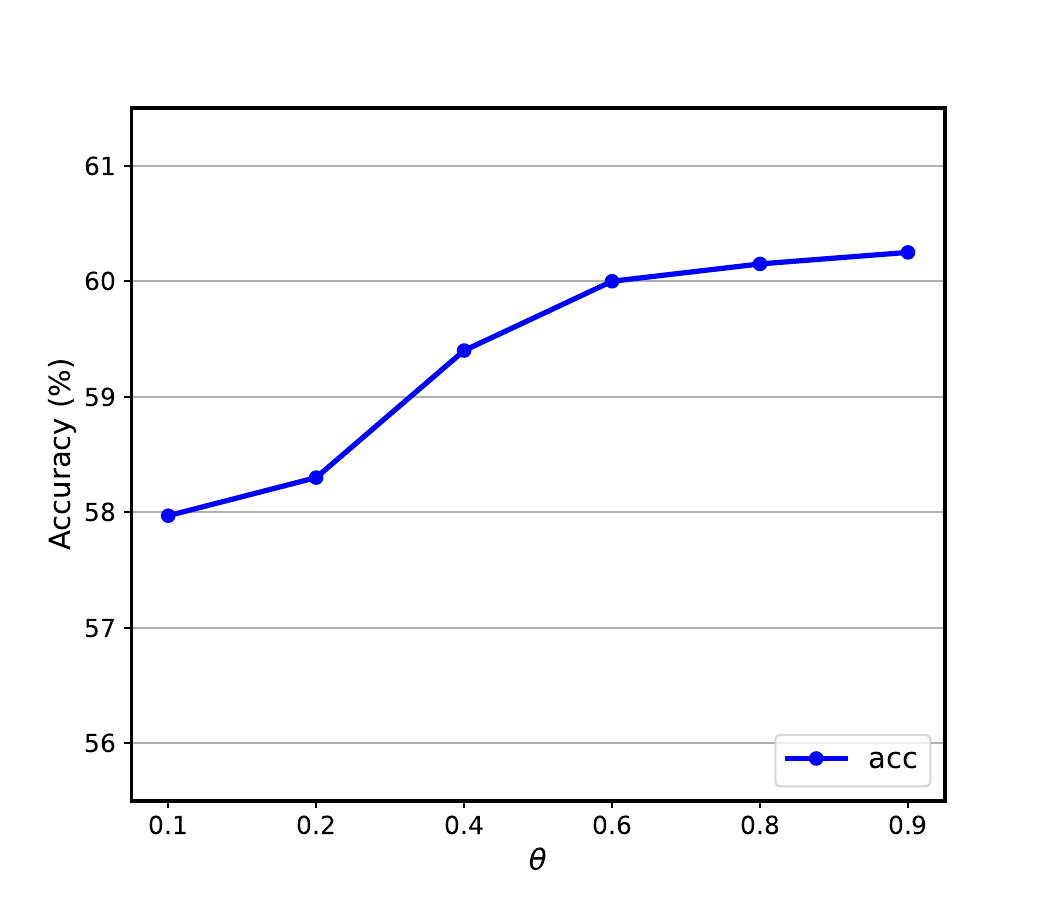}&
			\includegraphics[width=0.25\linewidth]{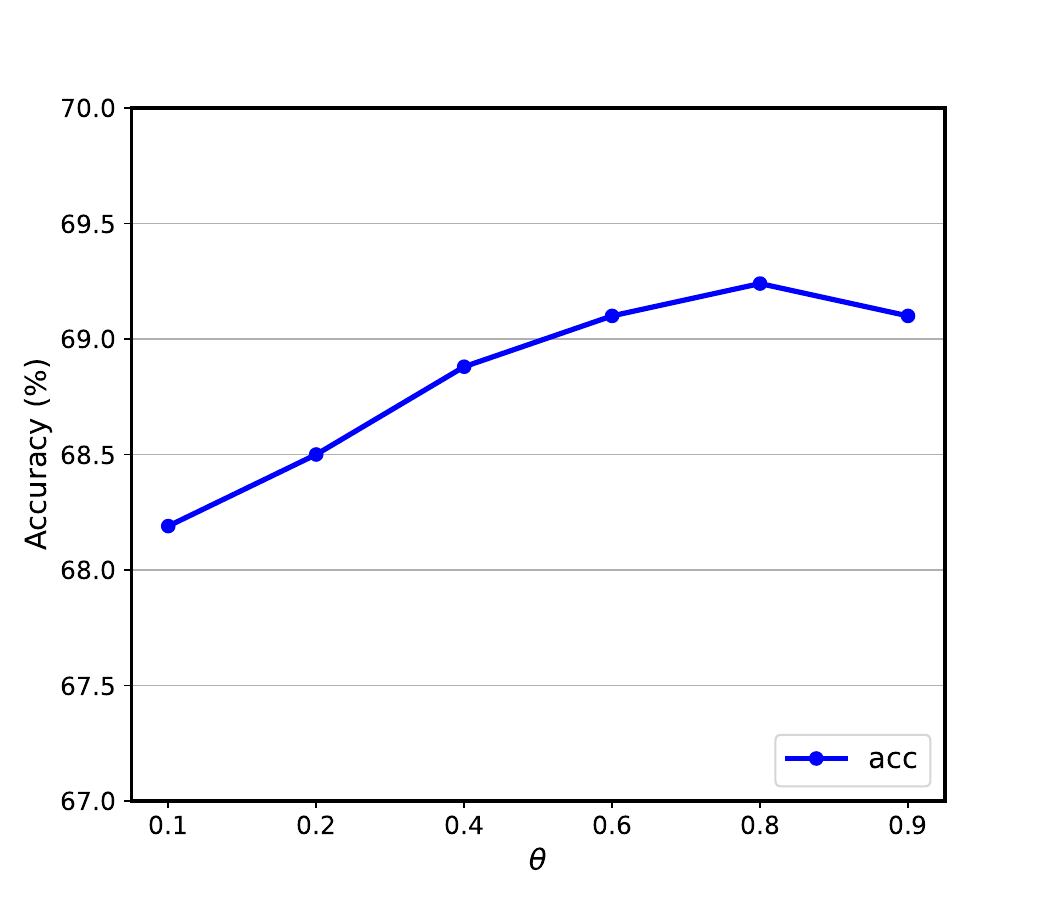}\\
			(a) Varying $\alpha$ on CUB& (b) Varying $\alpha$ on ImageNet& 
			(c) Varying $\theta$ on CUB& (d) Varying $\theta$ on ImageNet\\
		\end{tabular}\vspace{-1mm}
		\caption{Hyper-parameters analysis. We show the ZSL performance variations on CUB and ImageNet by adjusting the value of cropping scale $\alpha$ in (a, b), the value of hybrid coefficient $\theta$ in (c, d).}
		\label{fig:hyper-para}
	\end{center}
\end{figure*}

\subsection{Qualitative Analysis}
\noindent\textbf{Interpretable Class Prediction of  LaZSL.} As shown in Fig. \ref{fig:visualization1}, we show the classifier of our LaZSL on various samples. For example,  LaZSL recognizes the first image as “Scissor Tailed Flycatcher”, as it formulates the classifier with attributes “\textit{black wings with white wingbars}”, “\textit{black head with a white throat}”, “\textit{a long, forked tail}”, which are important factors of  “Scissor Tailed Flycatcher”. Different from standard CLIP which only computes a similarity between the whole image and prompts of class names, LaZSL can accurately predict classes according to the corresponding attributes. That is,  attributes are the interpretable factors for our method.

\vspace{3mm}
\noindent\textbf{Classification Comparison between LaZSL and DCLIP.} To intuitively show the effectiveness of local alignment in interpretable ZSL, we show the top-5 classification results of LaZSL and DCLIP \cite{Menon2022VisualCV}. Part results are shown in Fig. \ref{fig:visualization2}. We can find that LaZSL accurately predicts classes for images, while DCLIP fails.  Especially for the challenging fine-grained images,  there is little difference between various fine-grained classes, e.g., “Brass Memorial Plaque” and “Pedestal”. Because LaZSL effectively aligns the fine-grained visual regions with their corresponding attributes via OT, which formulates a strong classifier for class prediction.

\subsection{Hyper-Parameter Analysis}
\noindent\textbf{Cropping Scale $\alpha$.} We study the cropping scale $\alpha$ to determine its effectiveness on our LaZSL. As can be seen from Fig. \ref{fig:hyper-para}(a)(b), LaZSL is not sensitive to  $\alpha$, and it achieves best performances on all datasets when $\alpha$ is set to 0.6. Additionally, when $\alpha$ is set to too large (e.g., $\alpha\geq0.6 $), the performance of  LaZSL decreases as it cannot conduct accurate local vision regions with attributes.  Accordingly, we experimentally set the $\alpha=0.6$ for all datasets. 

\noindent\textbf{Hybrid Coefficient $\theta$.} We study the hybrid coefficient $\theta$ to determine its effectiveness on our LaZSL. As illustrated in Fig. \ref{fig:hyper-para}(c)(d), the performance of LaZSL improves as local visual information is progressively integrated into visual-semantic alignment and class prediction. This indicates that local alignment is crucial for interpretable ZSL. However, if $\theta$ is set too high (e.g., $\theta \geq 0.9$), performance may degrade. This is because excessive local information can lead to knowledge forgetting in the visual encoder of CLIP. Based on our experiments, we set $\theta = 0.8$ to fuse key local visual features with a few global features in LaZSL.

\subsection{Cost Analysis}
In fact, we follow DCLIP \cite{Menon2022VisualCV} to take GPT-3 to generate the attribute sets for each
class, which is similar to CuPL. As such, the cost of attribute generation using LLM is consistent for these methods, and the additional costs of LaZSL come from the OT optimization. The cost statistics are presented in Table \ref{table:cost}. Results show that our LaZSL does not take too much additional time and memory costs.
\begin{table}[htbp]
	\small
	\centering
	\caption{ Costs of various methods for handling one sample.} \label{table:cost}\vspace{-3mm}
	{
		\begin{tabular}{l|cc}
			
			\hline
			Method&Time Cost (s) &Memory Cost (MB)\\
			\hline
			CuPL \cite{PrattCLF23}                     & 0.015&619.62\\
			DCLIP  \cite{Menon2022VisualCV}               & 0.015&612.87\\
			LaZSL                  & 0.070&769.01\\
			\hline
		\end{tabular}
	}\vspace{-3mm}
\end{table}

\section{Conclusion}
\label{sec5}
In this paper, we address the visual-semantic alignment challenge in interpretable VLM-based ZSL. To this end, we propose LaZSL, which leverages local visual-semantic alignment through Optimal Transport to effectively align fine-grained visual information with semantic attributes. This enables LaZSL to learn an accurate and interpretable classifier for ZSL, achieving strong performance and domain generalization. Notably, in contrast to most existing VLM-based ZSL methods that require additional training for model fine-tuning or prompt adjustment, LaZSL enhances the ZSL performance of CLIP without any extra training. Both qualitative and quantitative results demonstrate the effectiveness of LaZSL.

\noindent \textbf{Limitations.} Indeed, our LaZSL approach depends on the quality of the attribute set generated by LLMs, which may occasionally produce irrelevant attribute descriptions that do not align well with the corresponding classes. Future work could focus on: i) designing improved prompts to guide LLMs in generating higher-quality attributes, and ii) developing filtering mechanisms to refine the generated attributes, ensuring that only the most relevant details are retained.

{
    \small
    \bibliographystyle{ieeenat_fullname}
    \bibliography{LAZSL}
}

\end{document}